\pdfoutput=1

\documentclass[11pt]{article}

\usepackage{emnlp2021}

\usepackage{times}
\usepackage{latexsym}

\usepackage{amsmath}
\usepackage{multirow}
\usepackage[utf8]{inputenc}
\usepackage{graphicx}
\usepackage{amssymb}
\usepackage{booktabs} 
\usepackage{verbatim} 

\usepackage{floatrow}
\usepackage{algorithm}
\usepackage{algpseudocode}
\usepackage{multicol}
\usepackage{xcolor}
\DeclareFloatFont{scriptsize}{\scriptsize}
\usepackage[T1]{fontenc}

\usepackage[utf8]{inputenc}

\usepackage{microtype}
\usepackage[bottom]{footmisc}
\title{Efficient Domain Adaptation of Language Models via Adaptive Tokenization}

\author{Vin Sachidananda\thanks{\hspace{1mm} Work done during an internship at Amazon.} \\
  Stanford University \\
  \texttt{vsachi@stanford.edu} \\\And
  Jason S. Kessler \\
  Amazon \\
  \texttt{jasokess@amazon.com} \\
  \And
  Yi-An Lai \\
  AWS AI HLT \\
  \texttt{yianl@amazon.com} \\
  }

\date{}

\begin{document}
\maketitle
\begin{abstract}
Contextual embedding-based language models trained on large data sets, such as BERT and RoBERTa, provide strong performance across a wide range of tasks and are ubiquitous in modern NLP. It has been observed that fine-tuning these models on tasks involving data from domains different from that on which they were pretrained can lead to suboptimal performance. Recent work has explored approaches to adapt pretrained language models to new domains by incorporating additional pretraining using domain-specific corpora and task data. We propose an alternative approach for transferring pretrained language models to new domains by adapting their tokenizers. We show that domain-specific subword sequences can be efficiently determined directly from divergences in the conditional token distributions of the base and domain-specific corpora. In datasets from four disparate domains, we find adaptive tokenization on a pretrained RoBERTa model provides $>$97\% of the performance benefits of domain specific pretraining. Our approach produces smaller models and less training and inference time than other approaches using tokenizer augmentation. While adaptive tokenization incurs a 6\% increase in model parameters in our experimentation, due to the introduction of 10k new domain-specific tokens, our approach, using 64 vCPUs, is 72x faster than further pretraining the language model on domain-specific corpora on 8 TPUs. 
\end{abstract}

\section{Introduction}
Pretrained language models (PLMs) trained on large ``base'' corpora, oftentimes $>$100GB of uncompressed text \citet{roberta,gpt3}, are used in many NLP tasks. These models first learn contextual representations in an unsupervised manner by minimizing a masked language modeling objective over a base corpus. This stage of unsupervised language model training is referred to as "pretraining". Subsequently, for supervised classification tasks, the output head of this pretrained model is swapped for a lightweight classifier and trained further on a classification objective over labeled data, referred to as ``fine-tuning''. 

Recent work has examined the transferability of PLMs \citet{dontstop} and their contextual representations to domains differing from their base corpora. On text classification tasks from four different domains, it was shown that continuing to pretrain RoBERTa's contextual embeddings on additional domain (DAPT) and/or task-specific data (TAPT) resulted in performance gains over only fine-tuning a baseline RoBERTa model. These performance gains, however, were inferior to each task's start-of-the-art metrics which were largely based on training versions of RoBERTa, or other LMs, from scratch on a large sample of in-domain data.

These performance gains come at substantial financial, time, and environmental costs in the form of increased computation, with pretraining an LM from scratch being the most expensive, using additional pretraining in the middle, and only fine-turning an off-the-shelf model the most economical.  

One observed advantage \citet{pubmedbert} that pretraining from scratch on in-domain data has over continual pretraining is that the tokenizer's vocabulary captures domain-specific terms. This allows semantics of those terms to be directly learned in their fixed embeddings, and relieves the language model from having to encode these semantics through the contextual embeddings of these domain-specific term's subwords. Recent work \citet{zhang-etal-2020-multi-stage,poerner-etal-2020-inexpensive} has shown adding whole words common to the target domain but absent from a PLM's tokenizer improves performance on single tasks. In this work, we show that augmenting an PLM with statistically derived subword tokens selected for domain association with simple embedding initializations and no further pretraining provide an effective means of adapting a PLM across tasks and domains. In contrast, both  \citet{zhang-etal-2020-multi-stage} and \citet{poerner-etal-2020-inexpensive} add inefficiencies by respectively requiring further masked language model (MLM) pretraining and doubling the resources needed for inference.

In this paper, we efficiently adapt a PLM by simply augmenting its vocabulary with domain-specific token sequences. We find that this adaptation, which requires no further pretraining, rivals the accuracy of domain and task-adapted pretraining approaches proposed in \citet{dontstop} but requires only a small fraction of the compute cost. 


\section{Related work}
\label{sec:rw}
\citet{dontstop} describes two complementary methods using a task's training data or a separate unlabeled domain-specific corpus to further pretrain an LM, denoted as Task-Adaptive Pretraining (TAPT) and Domain-Adaptive Pretraining (DAPT) respectively. This paper shows the value of employing additional in-domain data in pretraining on four domains relative to only fine-tuning a PLM. Our approach is directly comparable to DAPT, as we only use in-domain corpora for adaptation. 

\citet{zhang-etal-2020-multi-stage} augment RoBERTa's vocabulary with in-domain OOV whole words. The most frequently occurring whole words are added until the OOV rate drops to 5\% on the task corpus. They randomly initialize weights and pretrain a model. This improves performance on TechQA and AskUbuntu.  \citet{tai-etal-2020-exbert} also augmented BERT with tokens selected by frequency (12k OOV wordpieces were used) and pretrained a modified version of BERT which allowed for only new token's embeddings to be modified while the original embeddings remained fixed. They found that using more than 12k augmented tokens didn't improve their biomed NER and relation extraction performance, and that, once augmented, performance improved with more pretraining (4-24 hours were studied.) 

\citet{poerner-etal-2020-inexpensive} augment BERT's vocabulary with all in-domain OOV whole words, adding ~31K tokens to bert-base-cased's ~29K wordpieces. They trained a word2vec model on an in-domain corpus and fit a linear transformation to project the word embeddings into the model's input embedding space. No further pretraining is done, but during finetuning, the original tokenizer and the adapted tokenizer are both used. For inference, the finetuned model is run with both the original tokenizer and the adapted tokenizer and the outputs are averaged. Their F1 score outperforms BERT on all eight biomedical NER tasks studied. The approach has the disadvantage of increasing the parameter size of bert-base-cased by 2.2x due to the embeddings of added tokens and doubles the resources needed for inference. 

\citet{superbizarre} demonstrates how Wordpiece tokenization does not capture the semantics of derivationally complex words as well as an approach using a modified version of Wordpiece designed to produce subword segmentations consisting of linguistic prefixes, suffixes and affixes \citet{dagobert}. This subword tokenizer outperformed WordPiece in determining words' polarity or their source domains. Experiments were conducted on novel embedding tokens in BERT via approaches including a projection-based method and mean pooling (both similar to \S \ref{sec:embed}).

Training language models from scratch in the domain of interest has been shown to provide improved in-domain performance when compared to out-of-domain PLMs \citet{clinicalbert}. In addition to \citet{dontstop}, prior work has shown the effectiveness of continued pretraining for domain adaptation of PLMs \citet{alsentzer-publicly, chakrabarty-imho, lee-biobert}. For the task of Aspect-Target Sentiment Classification, \citet{rietzler-etal-2020-adapt} uses both DAPT and task-specific fine-tuning in order to adapt language models representations. Identifying domain-characteristic words is a well-studied problem, and many metrics have been proposed for this task through comparing the distributions of tokens in contrasting corpora \citet{keyness,mcq,kessler-2017-scattertext}. \citet{muthukrishnan-etal-2008-detecting} used the pointwise KL-divergence to distinguish informativeness of key phrase candidates in a domain corpus relative to a background. 


\section{Adaptive tokenization of contextual embeddings}

We define adaptive tokenization (AT) as the process of augmenting a PLM's tokenizer and fixed subword embeddings with new entries taken from a novel corpus. AT consists of two goals which must be achieved for domain adaptation. First,  selection of domain-specific tokens, with which to augment a pretrained tokenizer, from an in-domain corpus must be determined. Second, an appropriate initialization in the input space of the contextual embedding models needs to be determined for additions to the tokenizer vocabulary. In this section, we detail approaches for each of these linked tasks.

\subsection{Tokenizer vocabulary augmentation}
\label{sec:vocab}
\noindent In this section, we detail approaches for identifying domain-specific token sequences to be added during tokenizer augmentation. Common tokenization schemes such as Byte Pair Encoding \citet{bpe} and WordPiece \citet{wordpiece,wordpiece2} are greedy algorithms and, as a result, merge subwords into individual tokens if such a sequence occurs with high relative frequency. When adapting a tokenizer our goal is to identify subword sequences which occur with high relative frequency in a domain specific corpus compared to the pretraining corpus. In Table \ref{tab:data}, we provide the corpora for each domain in which experimentation is conducted. Next, we show how to operationalize this framework to find domain-specific token sequences.

\subsection{Identifying domain-specific token sequences}
In this section, we detail our approach for selection of token sequences which are both difficult to represent in a base tokenizer and have large disparities in occurrence between domain-specific and base corpora. Conceptually, we would like to add new tokens to the source tokenizer which are sequences of existing tokens and, in the in-domain corpus, are extensions of existing token sequences.



\begin{algorithm*}[h]
  \centering
\caption{Selection of Domain-Specific Token Sequences for Tokenizer Augmentation}\label{alg:selection}
\begin{algorithmic}
\Require {$\text{Base Tokenizer} \hspace{2mm} Tok, \text{Base LM} \hspace{2mm} LM_{base},$ $\text{Base and Domain Unigram Dists.} \hspace{2mm} U_{base}, U_{domain}$,  $\text{Base and Domain Seq. Dists.} \hspace{2mm} T_{base} \text{= \{\}},  T_{domain} \text{= \{\}}$ $\text{Min. Seq. Frequency} \hspace{2mm} F_{min}, \text{\# Aug. to make} \hspace{2mm} N,$ $\text{Max Aug. Length} \hspace{2mm} L,$ $\text{Augmentations} \hspace{2mm} = []$}

\State{\textbf{(I) Computing Empirical Token Sequence Distributions}}

\For{word, count $(w, count)$ in $U_{base}$} \Comment{Do the same for Domain Corpus}
    \State $Seq [t_0, t_1, ..., t_n] := Tok(w)$ 
    \For{i in [1,n]}
        \State $T_{base}[Seq[:i]] \mathrel{{+}{=}} count$
    \EndFor
\EndFor

\State $T_{domain}\text{.values()} \mathrel{{/}{=}} \text{sum}(U_{domain}\text{.values()} )$ \Comment{Normalize Sequence Distributions}
\State $T_{base}\text{.values()} \mathrel{{/}{=}} \text{sum}(U_{base}\text{.values()} )$

\State{\textbf{(II) Domain shift scoring of Token Seq. Dists. with Conditional KL Divergence}}
\State $Score_{DKL} = \{\}$
\For{Seq in $T_{base} \bigcap T_{domain}$}
    \State $Score_{DKL}[Seq] := T_{domain}[Seq] * \log{\frac{T_{domain}[Seq]}{T_{base}[Seq]}}$ 
\EndFor

\State{\textbf{(III) Selection of Token Sequences for Augmentation}}

\State $\text{SortDescending} (Score_{DKL}) $

\For{Seq in $Score_{DKL}$}
    \If {$\text{Len}(\text{Augmentations}) = N$}
        \State{\textbf{break}}
    \EndIf
    \If{$\text{Len}(Seq) < L$ $\textbf{AND}$ $T_{domain} > F_{min}$ $\textbf{AND}$ $T_{base} > F_{min}$}
        \State{\text{Augmentations.append(Seq)}}
    \EndIf
\EndFor
\State{\textbf{return} $\text{Augmentations}$}
\end{algorithmic}
\end{algorithm*}

\noindent \textbf{(I) Computing Empirical Token Sequence Distributions} We first compute counts of sequences of $[1,\lambda]$ subword tokens ($s$) in each corpus $C$, namely the source corpus for RoBERTa ($S$) and the in-domain corpus which is the target of our adaptation ($D$). The source language model's tokenizer (namely Roberta-base) is used as the source of subword tokens. The counts of each subtoken sequences are represented as $C_s$, where $C$ is the corpus and $s$s is the subword sequence. If $s$ does not appear in $C$, $C_s = 0$.  We only retain sequences occurring at least $\phi=20$ times in one corpus. The maximum subword token sequence length ($\lambda$) is $10$. We limit subtoken sequences to word boundaries as detected through whitespace tokenization.

Next, we predict how ``phrase-like'' a sequence of tokens $C_s$  is, using a probability $P_C(s)$. Define 
$$P_C(s) = \frac{C_s}{C_t}$$
where $t$ is first $|s|-1$ subtoken sequence of $s$. These probabilities should be thought of as the surprise of the sequence $s$ in the corpus being counted and are indicative of the how phrase-like $s$ is. 

As an example, consider a hypothetical corpus consisting of documents written about classical music. Roberta-base's tokenizer splits ``oboe'' into the subtokens $\langle \textnormal{ob}, \textnormal{oe} \rangle$.  In this classical music corpus, the portion of tokens following ``ob'' which are ``oe'' (composing in the word ``oboe'') is surely much higher than in a general base corpus where other words staring with the ``ob'' subtoken like ``obama'' (tokenized as $\langle \textnormal{ob}, \textnormal{ama} \rangle$) are much more frequent and ``oboe'' much less.
 
\noindent \textbf{(II) Domain shift scoring of Token Sequence Distributions with Conditional KL Divergence} In order to characterize these differences in probabilities, we use the pointwise KL-divergence. Letting $p$ and $q$ be probabilities, the pointwise KL-divergence is defined as:

$$D_{KL}(p \parallel q)) = p \log \frac{p}{q}$$

Let the sequence relevance score $R(s)$ be defined as

$$R(s) = D_{KL}(P_D(s) \parallel P_S(s)).$$

$R(s)$ indicates how much the phrase-like probability of sequence $s$ in the in-domain corpus $D$ ($P_D(s)$) diverges from the baseline phrase-like probability of $s$ in the base corpus $S$.   

\noindent \textbf{(III) Selection of Token Sequences for Tokenizer Augmentation}
For all experiments, we add the $\eta=10K$ sequences with the largest $R$, sorted irrespective of sequence length, to the domain-augmented tokenizer. 

This introduces of 7.68M parameters (embedding size $768\times10$K new tokens), a 6\% increase over Roberta-base's 125M.\footnote{github.com/pytorch/fairseq/tree/master/examples/roberta}


\subsection{Initialization approaches for AT}
\label{sec:embed}
In this section, we provide two approaches to impute contextual embedding input representations for tokens added in \S \ref{sec:vocab}.  

\noindent \textbf{Subword-based initialization} In this common initialization \citet{casanueva-etal-2020-efficient,Vuli2020ProbingPL,superbizarre}, additions to the tokenizer are embedded as the mean of their Roberta-base fixed subword embeddings. In cases where all a novel word's subwords are unrelated to its specific, in-domain meaning, this initialization may cause unwanted model drift in fine-tuning for unrelated tokens with similar fixed embeddings. 

\begin{algorithm}
\caption{Projection-Based Initialization of Augmented Tokens}\label{alg:augment_algorithm}
\begin{algorithmic}
\Require $\text{LM Input Embeddings} \hspace{2mm} C_s, \text{Base and}$ $\text{Domain Learned Input Embeddings}$ $X_s, X_t$, and Embedding Size $d$.

\State{\textbf{(I) Learn Mapping $\hat{\mathcal{M}}$: $C_s \rightarrow X_s$ with SGD}:}
\State $\hat{\mathcal{M}} = \arg \min_{\mathcal{M} \in \mathbb{R}^{d \times d}} \| \mathcal{M} X_s - C_s \|_F$


\State{\textbf{(II) Get Inits. for Aug. Tokens using $\hat{\mathcal{M}}$}:}
\State $C_t = \hat{\mathcal{M}}X_t$
\State \Return $C_t$
\end{algorithmic}
\end{algorithm}

\begin{table*}[h!]
  \centering
    \scalebox{1.1}{
\begin{tabular}{lllccccc} \toprule
    Domain & Pretrain Corpus [\# Tokens] & Task & Task Type & Train (Lab.) & Dev. & Test & Classes  \\ \midrule
    \multirow{ 2}{*}{BioMed} & \multirow{ 2}{*}{1.8M papers from S2ORC [5.1B]} & ChemProt & relation classification & 4169 & 2427 & 3469 & 13  \\
    & & RCT &  abstract sent. roles & 18040 & 30212 & 30135 & 5    \\ \midrule
    \multirow{ 2}{*}{CS} & \multirow{ 2}{*}{580K papers from S2ORC [2.1B]}  & ACL-ARC & citation intent & 1688 & 114 & 139 & 6  \\
    & & SciERC & relation classification & 3219 & 455 & 974 & 7    \\ \midrule
    News & 11.9M articles [6.7B] & HyperPartisan & partisanship & 515 & 65 & 65 & 2   \\ \midrule
    Reviews & 24.75M Amazon reviews [2.1B] & IMDB & review sentiment & 20000 & 5000 & 25000 & 2 \\ 
    \bottomrule
\end{tabular}
} \vspace*{-2mm}
  \caption{Specifications of the various target task and pretraining datasets to replicate experiments in \citet{dontstop}.  Due to the restrictions on accessible papers in S2ORC, we are using versions of BioMed and CS which are approximately 33\% and 74\% smaller than were used in \citet{dontstop}. Sources:  S2ORC \citet{s2orc}, News \citet{fakenews}, Amazon reviews \citet{amznrev}, CHEMPROT \citet{chemprot}, RCT \citet{rct}, ACL-ARC \citet{aclarc}, SCIERC \citet{scierc}, HYPERPARTISAN \citet{hyperpartisan}, and IMDB \citet{imdb}.}
  \label{tab:data}
\end{table*}

\noindent \textbf{Projection-based initialization} To mitigate possible issues with averaging subword embeddings, we also consider projections between static token embeddings to the input space of contextual embeddings, similar to \citet{poerner-etal-2020-inexpensive}. 

To summarize this approach, our goal is to learn a mapping between the input token embeddings in RoBERTa, $C_{base}$, and word2vec token embeddings learned independently on the base\footnote{See \S \ref{impdet} for how the RoBERTa source corpora is approximated to form our base corpus.} and domain specific corpora, $X_{base}, X_{domain}$. The tokens in $C_{base}$ include the original RoBERTa tokens while those in $X_{base}$ and $X_{domain}$ include both the original RoBERTa tokens and the augmented tokens found using adaptive tokenization detailed in \S 3.2. First, a mapping $M$, parametrized as a single layer fully connected network, from $X_{base}$ to $C_{base}$ is learned which minimizes distances, on the original set of tokens in RoBERTa. The goal of this mapping is to learn a function which can translate word2vec token embeddings to the input space of RoBERTa. Then, the learned mapping $M$ is applied to $X_{domain}$ in order to obtain initializations in the input space of RoBERTa for the augmented tokens found using the approach in \S 3.2. The operations involved in this approach are detailed in Algorithm 2.

\begin{table*}[t]
  \centering
  \begin{tabular}{llc|ccc|cc||c}  \toprule
    Domain & Task & RoBERTa & DAPT & TAPT & DAPT + TAPT & AT (Mean) & AT (Proj) & State-of-the-art (in 2020) \\ \midrule
    \multirow{ 2}{*}{BioMed$^{*}$} & ChemProt & 81.9$_{1.0}$ & \underline{84.2$_{0.2}$} & 82.6$_{0.4}$ & \textbf{84.4$_{0.4}$} &  83.6$_{0.4}$ & 83.1$_{0.3}$ & 84.6 \\
     & RCT & 87.2$_{0.1}$ & \underline{87.6$_{0.1}$} & 87.7$_{0.1}$ & \textbf{87.8$_{0.1}$} & 87.5$_{0.4}$ & \underline{87.6$_{0.3}$} & 92.9  \\ \midrule
    \multirow{ 2}{*}{CS$^{*}$} & ACL-ARC & 63.0$_{5.8}$ & \underline{75.4$_{2.5}$} & 67.4$_{1.8}$ & \textbf{75.6$_{3.8}$} & 70.1$_{2.0}$ & 68.9$_{1.6}$ & 71.0\\
     & SciERC & 77.3$_{1.9}$ & 80.8$_{1.5}$ & 79.3$_{1.5}$ & 81.3$_{1.8}$ & \underline{\textbf{81.4}$_{0.4}$} & 81.2$_{1.2}$ & 81.8 \\ \midrule
    News & HyperPartisan & 86.6$_{0.9}$ & 88.2$_{5.9}$ & 90.4$_{5.2}$ & 90.0$_{6.6}$ & \underline{\textbf{93.1}$_{4.2}$} & 91.6$_{5.5}$ &  94.8 \\ \midrule
     Reviews & IMDB & 95.0$_{0.2}$ & 95.4$_{0.1}$ & 95.5$_{0.1}$ & \textbf{95.6}$_{0.1}$ & 95.4$_{0.1}$ & \underline{95.5$_{0.1}$} & 96.2 \\
     \bottomrule
  \end{tabular}\vspace*{-1mm}
  \caption{Results of different adaptive pretraining methods compared to the baseline RoBERTa. AT with mean subword and projective initializations are denoted as AT (Mean) and AT (Proj) respectively. Stddevs are from 5 seeds. Results for DAPT, TAPT, DAPT+TAPT, and state-of-the-arts are quoted from \citet{dontstop}.  The highest non-state-of-the-art result is bolded, since the state-of-the-art functions as a performance ceiling, leveraging both domain-specific pretraining and an adapted tokenizer. The best of the three approaches which utilize only source and domain domain data before fine-tuning (i.e., DAPT and AT) is underlined.  *Due to restrictions on accessible papers in S2ORC, The BioMed and CS pretraining corpora used were respectively 33\% and 74\% smaller than the versions in \citet{dontstop}. Note that state-of-the-art numbers are current at the time of \citet{dontstop}, and are from the following works: ChemProt: S2ORC-BERT \citet{s2orc}, RCT: Sequential Sentence Classification \citet{rct_sota}, ACL-ARC: SciBert \citet{arc_sota}, SciERC: S2ORC-BERT \citet{s2orc}, HyperPartisan: Longformer \citet{longformer}, IMDB: XLNet Large \citet{xlnet-large}.}
\end{table*}

\begin{table}[t]
  \centering
  \small
\begin{tabular}{lllccccc} \toprule
    Method & Hardware Specs. & Runtime [h:m:s]  \\ \midrule
    DAPT & 8x TPU V-3 & 94 hours  \\
    AT (Mean) & 64x vCPUs & 1:17:35  \\
    AT (Projection) & 64x vCPUs & 4:54:58  \\ 
    \bottomrule
\end{tabular}\vspace*{-1mm}
  \caption{Runtime and hardware specifications for AT compared to DAPT. The vast majority of the time is spent reading the corpus and creating token distributions.  Runtimes are based on the CS 8.1B token corpus. The DAPT runtime is mentioned in Github Issue 16 in \citet{dontstop} and the AT runtimes are linearly extrapolated (an overestimate) from our observed runtime on the open version of CS, a 2.1B token corpus.  We needed to perform this extrapolation since the full CS corpus which was used to benchmark \citet{dontstop} is unavailable in S2ORC. ``64x vCPUs'' indicate the equivalent of an AWS ml.m5.16xlarge EC2 instance was used to determine which subtoken sequences to use for vocabulary augmentation and compute their embeddings. The times reported for AT (Mean) and AT (Projection) where from a single run, with precomputed base corpus token counts and embeddings.}
  \label{tab:perf}
\end{table}
\section{Experimentation}

\begin{table*}[h!]
  \centering
  \scalebox{1.15}{
\begin{tabular}{llll} \toprule
    BioMed & CS & News & Reviews  \\ \midrule
    \text{[inc, ub, ated]} $\rightarrow$ \text{incubated} & \text{[The, orem]} $\rightarrow$ \text{Theorem} & \text{[t, uesday]} $\rightarrow$ \text{tuesday} & \text{[it, 's]} $\rightarrow$ \text{it's}  \\
    \text{[trans, fect]} $\rightarrow$ \text{transfect} & \text{[L, em, ma]} $\rightarrow$ \text{Lemma} & \text{[ob, ama]} $\rightarrow$ \text{obama} & \text{[that, 's]} $\rightarrow$ \text{that's} \\
    \text{[ph, osph, ory]} $\rightarrow$ \text{phosphory} & \text{[vert, ices]} $\rightarrow$ \text{vertices} & \text{[re, uters]} $\rightarrow$ \text{reuters} & \text{[sh, oes]} $\rightarrow$ \text{shoes} \\
    \text{[mi, R]} $\rightarrow$ \text{miR} & \text{[E, q]} $\rightarrow$ \text{Eq}  & \text{[iph, one]} $\rightarrow$ \text{iphone} & \text{[doesn, 't]} $\rightarrow$ \text{doesn't} \\
    \text{[st, aining]} $\rightarrow$ \text{staining} & \text{[cl, ust, ering]} $\rightarrow$ \text{clustering} & \text{[ny, se]} $\rightarrow$ \text{nyse} & \text{[didn, 't]} $\rightarrow$ \text{didn't} \\
    \text{[ap, opt, osis]} $\rightarrow$ \text{apoptosis} & \text{[H, ence]} $\rightarrow$ \text{Hence} & \text{[get, ty]} $\rightarrow$ \text{getty} & \text{[can, 't]} $\rightarrow$ \text{can't} \\
    \text{[G, FP]} $\rightarrow$ \text{GFP} & \text{[Seg, mentation]} $\rightarrow$ \text{Segmentation} & \text{[inst, agram]} $\rightarrow$ \text{instagram} & \text{[I, 've]} $\rightarrow$ \text{I've} \\
    \text{[pl, asm]} $\rightarrow$ \text{plasm} & \text{[class, ifier]} $\rightarrow$ \text{classifier} & \text{[bre, xit]} $\rightarrow$ \text{brexit} & \text{[b, ought]} $\rightarrow$ \text{bought} \\
    \text{[ass, ays]} $\rightarrow$ \text{assays} & \text{[Ga, ussian]} $\rightarrow$ \text{Gaussian} & \text{[nas, daq]} $\rightarrow$ \text{nasdaq} & \text{[you, 'll]} $\rightarrow$ \text{you'll}  \\
    \text{[ph, osph, ory, lation]} $\rightarrow$ \text{phosphorylation} & \text{[p, olyn]} $\rightarrow$ \text{polyn} & \text{[ce, o]} $\rightarrow$ \text{ceo} & \text{[kind, le]} $\rightarrow$ \text{kindle} \\
    \bottomrule
\end{tabular}
}\vspace*{-3mm}
  \caption{Samples of token sequences with large JSD between base and domain corpora sequence distributions; all of these sequences were added during AT to the Roberta-Base tokenizer.}
  \label{tab:vocab}
\end{table*}

In this section, we perform evaluation of our adaptation approach on six natural language processing tasks in four domains, BioMedical, Computer Science, News, and Reviews, following the evaluations in \citet{dontstop}. Due to resource constraints, we perform experimentation on all datasets in \citet{dontstop} excluding the Helpfulness dataset from the reviews domain and the Hyperpartisan dataset in the news domain. Each of the excluded datasets contain greater than 100K training examples, resulting in greater than 12 hours of time required for finetuning on 8 Tesla V100 GPUs for a single seed.  

\noindent \textbf{Approaches} Roberta-base, a commonly used PLM with high performance, is used as a baseline on which supervised finetuning is performed separately for each dataset. Additionally, we compare AT to the DAPT method from \citet{dontstop}. As we do not make use of task specific data (i.e., the training data used in fine-tuning), AT is comparable to DAPT in terms of the data utilized.  We focus on using large, in-domain data sets which are commonly used in further pretraining (rather than variably sized task-data) since their size both allows for reliable extraction of characteristic subtoken sequences to use in tokenizer augmentation. Adaptive tokenization for task-specific data is future work.

\noindent \textbf{Classification Architecture} We use the same classification architecture as in \citet{dontstop}, originally proposed in \citet{bert}, in which the final layer's [CLS] token representation is passed to a task-specific feed forward layer for prediction. All hyperaparameters used in experimentation are equivalent to either the "mini", "small", or "big" hyperparameter sets from \citet{dontstop}. 

\noindent \textbf{Results} We find that adaptive tokenization improves performance when compared to the baseline RoBERTa model in all four of the domains on which experimentation is performed. AT provides 97\%  of the aggregate relative improvement attained by DAPT respectively over Roberta-base while providing an order of magnitude efficiency gain detailed in Table \ref{tab:perf}. We do not see a significant difference in the performance of AT models based on the Mean or Proj initialization schemes. Given that Mean initialization required half the time as Proj, we recommend its use over Proj.

\section{Discussion}


\subsection{Resource Efficiency in LM Adaptation}

Current approaches for training and adapting LMs have resulted in negative environmental impact and high computational resource budgets for researchers. PLMs incur significant compute time during pretraining, typically requiring numerous days of training on $\geq8$ GPUs or TPUs \citet{roberta,bert,dontstop}. In Table \ref{tab:perf}, we provide a runtime comparison between continued pretraining and AT. We find that AT provides a $~$72x speedup compared to DAPT and does not require a GPU or TPU to run.  The most resource-intensive portion of this procedure involves indexing the corpora and conducting subtoken sequence counts.

In addition to time and resources, the environmental impact of pretraining BERT with a single set of hyperparameters incurs a carbon footprint of approximately 1.5K pounds of CO$_2$ emissions, more than the average monthly emissions of an individual \citet{strubell2019}. Continued pretraining, which has a similar resource budget to BERT, exacerbates this problem \citet{greenai}. Lastly, we find that the cloud computing costs associated with continual pretraining for both a single domain and set of hyperparameters are ~\$750 compared to around \$4.77 (using a ml.m5.16xlarge EC2 instance for 1:17) for AT on cloud computing platforms when using non-preemptible instances. High costs associated with the training of NLP models has led to inequity in the research community in favor of industry labs with large research budgets \citet{strubell2019}.

\subsection{Augmented Token Sequences selected in each domain}
\noindent In Table \ref{tab:vocab}, we provide examples of augmented vocabulary selected by our adaptive tokenization algorithm for each of the four domains used in experimentation. In each domain, the augmented tokens identified by AT correspond to domain-specific language. For instance, augmented tokens in the Reviews domain token sequences often contain contractions such as ``I've'' and ``it's'', which are frequently used in informal language. In the News domain, augmented tokens include financial terms such as ``NYSE'' and ``Nasdaq'' along with media outlets such as ``Reuters'' and ``Getty''. Many of the augmented tokens in the Computer Science domain are mathematical and computing terms such as ``Theorem'', ``Lemma'', ``Segmentation'', and ``Gaussian''. Lastly, augmented tokens in the BioMedical domain are largely concerned with biological mechanisms and medical procedures such as ``phosphorylation'', ``assays'', and ``transfect''.

\subsection{Future directions}
While we have evaluated this approach on Roberta-base, it can be used on any PLM which uses subword tokenization.  It would be interesting future work to see if the performance gain will hold on larger PLMs with richer vocabularies or on smaller PLMs. One may speculate the benefit of AT is due to encoding non-compositional subword tokens in the input embedding space. And furthermore, this lifts some of the responsibility for encoding their semantics from the LM's interior weights. Since these non-compositional tokens are characteristic to the domain corpus, their representations may be important to the end task and and need to be learned or improved during fine-tuning. If this is the case, then perhaps models with fewer interior weights benefit more from AT since the connection between the non-compositional tokens would be built into the input, allowing interior weights to better learn the semantics of novel non-compositional tokens and opposed to also having to learn the component tokens' connection. 

While this work tests AT on an English language PLM, it can hypothetically be applied to any PLM regardless of its source language(s). Exploring how AT can work with additional pretraining on domain data is clear future work. \citet{tai-etal-2020-exbert} show that specialized further pretraining on domain data on using a model augmented with domain characteristic whole word tokens results in an improved performance/pretraining time curve.  It would also be fruitful to explore how that curve changes when using more efficient pretraining techniques such as in \citet{clark2020electra}.

While we compared different novel token sequence embedding techniques, we did not study different ways of identifying subtoken sequences to add. Comparing AT to approaches such adding whole word tokens \citet{tai-etal-2020-exbert} would confirm our hypothesis that phrase-like token sequences are useful. 

Experimenting with the number of subtoken sequences added to the tokenizer ($\eta$ fixed at $10K$) may also be worthwhile. While \citet{tai-etal-2020-exbert} found $12K$ tokens additions optimal, \citet{poerner-etal-2020-inexpensive} added $310K$ tokens. Seeing the trade-off between added tokens and performance would be useful, as each additional parameter increases the model size.

Our approach requires new tokens to appear $\phi$ times in both the source and domain corpora.  While this was necessary in order to produce source-corpus word embeddings in Proj, it does not allow for domain-exclusive subtoken sequences to be added to the tokenizer.  Abandoning this requirement for Mean may lead to a better set of token augmentations.

We can also experiment with other subtoken candidate selection techniques. For example, \citet{Schwartz2013} used pointwise mutual information (PMI) to determine how phrase-like candidates word sequences were. PMI is the log ratio of the probability of a phrase vs. the product of the probability of its component unigrams. While our approach considers the probability of a subtoken given a preceding sequence, it, unlike PMI, does not consider the probability of that following subtoken in isolation. This may lead to domain-specific subtokens sneaking into augmented token sequences, such as the contraction tokens added to the reviews Reviews tokenizer in Table \ref{tab:vocab}.

\subsection{Implementation details}
\label{impdet}
The code is in preparation for release.  The hyperparameter search used was ROBERTA\_CLASSIFIER\_MINI from \citet{dontstop} from their codebase \url{https://github.com/allenai/dont-stop-pretraining}. Token counts for RoBERTa-base were estimated using English Wikipedia 20200501.en and an open source book corpus from \url{https://storage.googleapis.com/huggingface-nlp/datasets/bookcorpus/bookcorpus.tar.bz2}. Word2vec embeddings were computed with Gensim \cite{rehurek2011gensim}, using the following parameters:
\texttt{Word2Vec(..., size=768, window=5, min\_count=100, epochs=2, sample=1e-5)}

\section{Conclusion}
In this paper, we introduced adaptive tokenization (AT) a method for efficiently adapting pretrained language models utilizing subword tokenization to new domains. AT augments a PLM's tokenization vocabulary to include domain-specific token sequences. We provide two approaches for initializing augmented tokens: mean subword and projections from static subword embeddings. AT requires no further language model pretraining on domain-specific corpora, resulting in a 38x speedup over pretraining on the corpora without specialized hardware. Across four domains, AT provides >97\% of the performance improvement of further pretraining on domain-specific data over Roberta-base. This initial work suggests that adapting the subword tokenization scheme of PLMs is an effective means of transferring models to new domains. Future work entails hybrid approaches using both AT and small amounts of LM pretraining, alternative metrics for augmented token selection, improved initialization of augmented token representations, and the use of task data. 

\section*{Acknowledgements}
We thank Yi Zhang, William Headden, Max Harper, Chandni Singh, Anuj Ahluwalia, Sushant Sagar, Jay Patel, Sachin Hulyalkar, and the anonymous reviewers for their valuable feedback.

\section*{Ethics statement}
As mentioned in \S 5, pretrained language models incur significant costs with respect to time, computational resources and environmental impact. Continued domain specific pretraining, which has a similar resource budget to BERT, exacerbates this problem \citet{greenai}. In this work, we provide approaches for adapting pretrained language models to new domains with an approach, Adaptive Tokenization, which seeks to minimize costs associated with continued domain specific pretraining. It should be noted that we do not decrease the resource and environmental associated with pretraining, only the costs for domain adaptive pretraining which are nevertheless sizable (e.g. 32 TPU days for DAPT). 

Additionally, we find that the cloud computing costs associated with continued domain specific pretraining on a single domain and set of hyperparameters are around \$750 compared to around \$5 for AT on a cloud computing platform. High costs associated with the training of NLP models has led to inequity in the research community in favor of industry labs with large research budgets \citet{strubell2019}, a problem we seek to ameliorate.

This work does not address the high resource cost in fine-tuning PLMs. Risks associated with this paper are that this work may encourage the use of PLMs in more settings, such as domains with small amounts of data, and introduce potentially harmful inductive biases which have been found in many commonly used PLMs. 

We include statistics about the data sets used in Table \ref{tab:data}, these data sets were introduced in \citet{dontstop} and open source.

\clearpage 

\bibliography{anthology,custom}
\bibliographystyle{acl_natbib}

\clearpage

\appendix
\onecolumn
\end{document}